\documentclass{article} 
\usepackage{iclr2018_conference,times}
\usepackage[utf8]{inputenc}
\usepackage[T1]{fontenc}
\usepackage{amssymb}   
\usepackage{amsmath}   
\usepackage{amsfonts}  
\usepackage{amsthm}    
\usepackage{latexsym}  
\usepackage{graphicx}  
\usepackage{xcolor}    
\usepackage{hyperref}  
\usepackage{url}       
\usepackage{xspace}    
\usepackage{booktabs}  
\usepackage{multirow}  
\usepackage{multicol}  
\usepackage{hhline}    
\usepackage{ifthen}
\usepackage[noend]{algpseudocode}

\newcommand\refsec[1]{Section~\ref{sec:#1}}

\newcommand\reffig[1]{Figure~\ref{fig:#1}}

\newcommand\reftab[1]{Table~\ref{tab:#1}}
\newcommand\refapp[1]{Appendix~\ref{sec:#1}}

\ifthenelse{\isundefined{\definition}}{\newtheorem{definition}{Definition}}{}
\ifthenelse{\isundefined{\assumption}}{\newtheorem{assumption}{Assumption}}{}
\ifthenelse{\isundefined{\hypothesis}}{\newtheorem{hypothesis}{Hypothesis}}{}
\ifthenelse{\isundefined{\proposition}}{\newtheorem{proposition}{Proposition}}{}
\ifthenelse{\isundefined{\theorem}}{\newtheorem{theorem}{Theorem}}{}
\ifthenelse{\isundefined{\lemma}}{\newtheorem{lemma}{Lemma}}{}
\ifthenelse{\isundefined{\corollary}}{\newtheorem{corollary}{Corollary}}{}
\ifthenelse{\isundefined{\alg}}{\newtheorem{alg}{Algorithm}}{}
\ifthenelse{\isundefined{\example}}{\newtheorem{example}{Example}}{}
\newcommand{\E}{\ensuremath{\mathbb{E}}} 

\newcommand\D[1]{\tilde{#1}}    

\newcommand\piw{\pi_\mathrm{w}} 
\newcommand\pin{\pi_\mathrm{n}} 
\newcommand\domnet{\textsc{DOMnet}\xspace}

\title{Reinforcement Learning on Web Interfaces using Workflow-Guided Exploration}

\author{Evan Zheran Liu$^\dagger$\thanks{First three authors contributed equally},
Kelvin Guu$^\ddagger$\footnotemark[1],
Panupong Pasupat$^\dagger$\footnotemark[1],
Tianlin Shi$^\dagger$, Percy Liang$^\dagger$ \\
$^\dagger$Department of Computer Science, $^\ddagger$Department of Statistics\\
Stanford University, Stanford, CA 94305, USA \\
\texttt{\{evanliu,kguu,ppasupat,tianlins\}@stanford.edu,pliang@cs.stanford.edu}
}

\iclrfinalcopy 

\begin{document}

\maketitle

\begin{abstract}
Reinforcement learning (RL) agents improve through trial-and-error,
but when reward is sparse and the agent cannot discover successful action sequences,
learning stagnates.
This has been a notable problem in training deep RL
agents to perform web-based tasks, such as booking flights or
replying to emails, where a single mistake can ruin the entire sequence of actions.
A common remedy is to ``warm-start'' the agent by pre-training it to
mimic expert demonstrations, but this is prone to overfitting.
Instead, we propose to
\emph{constrain exploration} using demonstrations.
From each demonstration, we induce high-level
``workflows'' which constrain the allowable actions at each time step
to be similar to those in the demonstration
(e.g., ``Step 1: click on a textbox; Step 2: enter some text'').
Our exploration policy then learns to identify successful workflows and 
samples actions that satisfy these workflows. Workflows prune out bad exploration directions 
and accelerate the agent's ability to discover rewards.
We use our approach to train a novel neural policy designed to
handle the semi-structured nature of websites, and
evaluate on a suite of web tasks, including the recent World of Bits
benchmark. We achieve new state-of-the-art results, and show that \emph{workflow-guided
exploration} improves sample efficiency over behavioral cloning by more
than 100x.
\end{abstract}

\section{Introduction}

We are interested in training reinforcement learning (RL) agents
to use the Internet (e.g., to book flights or reply to emails) by directly
controlling a web browser. Such systems could expand the capabilities
of AI personal assistants \citep{stone2014amazon}, which are currently limited to interacting
with machine-readable APIs, rather than the much larger world of human-readable
web interfaces.

Reinforcement learning agents could learn to accomplish tasks using these
human-readable web interfaces through trial-and-error \citep{sutton1998reinforcement}.
But this learning process can be very slow in tasks with sparse reward,
where the vast majority of naive action sequences lead to no reward signal \citep{vecerik2017leveraging, nair2017overcoming}.
This is the case for many web tasks, which involve a large action space (the
agent can type or click anything) and require a well-coordinated sequence of actions to succeed.

A common countermeasure in RL is to pre-train the agent to mimic expert
demonstrations via behavioral cloning \citep{pomerleau1991efficient, kim2013learning}, encouraging it to take similar actions in
similar states. But in environments with diverse and complex states such as websites,
demonstrations may cover only a small slice of the state space, and it is
difficult to generalize beyond these states (overfitting). Indeed, previous work has found
that warm-starting with behavioral cloning often fails to improve over pure RL \citep{shi2017wob}.
At the same time, simple strategies to combat overfitting (e.g. using fewer
parameters or regularization) cripple the
policy's flexibility \citep{bitzer2010using}, which is required for complex spatial and structural reasoning in user interfaces.

In this work, we propose a different method for leveraging demonstrations.
Rather than training an agent to directly mimic them, we use demonstrations to
\emph{constrain exploration}. By pruning away bad exploration directions, we
can accelerate the agent's ability to discover sparse rewards. Furthermore,
because the agent
is not directly exposed to demonstrations, we are free to use a sophisticated
neural policy with a reduced risk of overfitting.

\begin{figure}\center
\begin{minipage}[c]{0.56\textwidth}
\includegraphics[width=\textwidth]{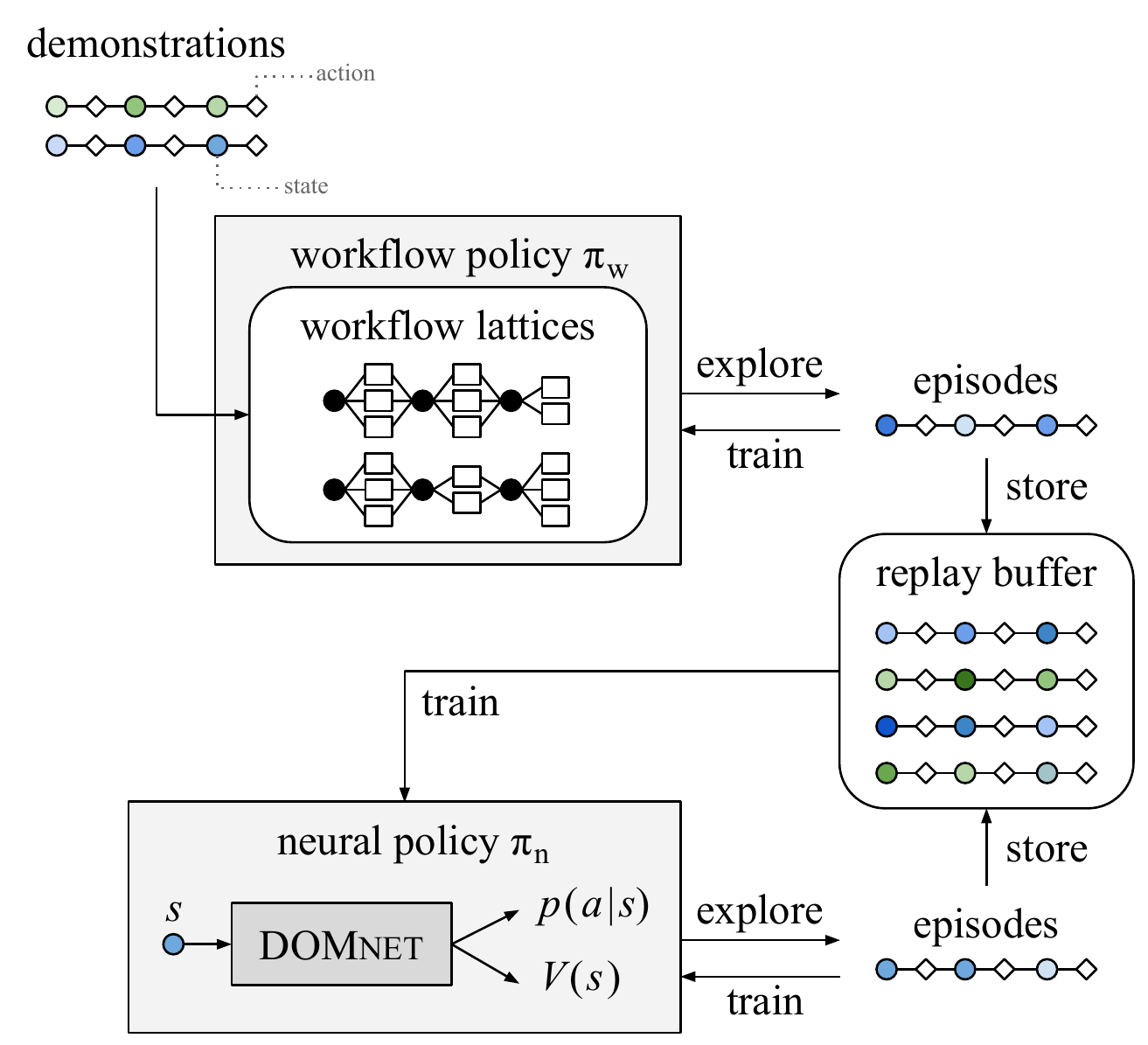}
\end{minipage}
\hspace{1.5em}
\begin{minipage}[c]{0.38\textwidth}\small
\textbf{Preprocessing:}
\begin{algorithmic}
\ForAll{demonstrations $d$}
  \State Induce workflow lattice from $d$
\EndFor
\end{algorithmic}
\vspace{.5em}
\textbf{Every iteration:}
\begin{algorithmic}
\State Observe an initial environment state
\State $\piw$ samples a workflow from a lattice 
\State Roll out an episode $e$ from the workflow
\State Use $e$ to update $\piw$
\If{$e$ gets reward $+1$}
  \State Add $e$ to replay buffer
\EndIf
\end{algorithmic}
\vspace{.5em}
\textbf{Periodically:}
\begin{algorithmic}
\If{replay buffer size > threshold}
  \State Sample episodes from replay buffer
  \State Update $\pin$ with sampled episodes
\EndIf
\State Observe an initial environment state
\State $\pin$ rolls out episode $e$
\State Update $\pin$ and critic $V$ with $e$
\If{$e$ gets reward $+1$}
  \State Add $e$ to replay buffer
\EndIf
\end{algorithmic}
\end{minipage}
\caption{
  \emph{Workflow-guided exploration (WGE).} After inducing workflow lattices from demonstrations,
the workflow policy $\piw$ performs exploration by sampling episodes from sampled workflows.
Successful episodes are saved to a replay buffer,
which is used to train the neural policy $\pin$.
}\label{fig:approach-overview}
\end{figure}

To constrain exploration, we employ the notion of a ``workflow'' \citep{deka2016erica}. For instance,
given an expert demonstration of how to forward an email, we might infer the
following workflow:
\begin{center}
Click an email title
$\to$ Click a ``Forward'' button\\
$\to$ Type an email address into a textbox
$\to$ Click a ``Send'' button
\end{center}

This workflow is more \emph{high-level} than an actual policy: it does not tell us exactly
which email to click or which textbox to type into, but it helpfully constrains
the set of actions at each time step. Furthermore, unlike a policy, it does not depend
on the environment state: it is just a sequence of steps that can be followed blindly.
In this sense, a workflow is \emph{environment-blind}.
The actual policy certainly should not be environment-blind,
but for exploration, we found environment-blindness to be a good inductive bias.

To leverage workflows, we propose the \emph{workflow-guided exploration} (WGE) framework
as illustrated in \reffig{approach-overview}:
\begin{enumerate}
\item For each demonstration, we extract a lattice of workflows that are
consistent with the actions observed in the demonstration (\refsec{workflows}).
\item We then define a \emph{workflow exploration policy} $\piw$ (\refsec{workflow-policy}),
which explores by first selecting a
workflow, and then sampling actions that fit the workflow. This policy
gradually learns which workflow to select through 
reinforcement learning.
\item Reward-earning episodes discovered during exploration enter a replay buffer,
which we use to train a more powerful and expressive neural network policy $\pin$
(\refsec{neural-policy}).
\end{enumerate}

A key difference between the web and traditional RL domains such as robotics
\citep{atkeson1997robot} or game-playing \citep{bellemare2013arcade} is that the state space involves a mix of
structured (e.g. HTML) and unstructured inputs (e.g. natural language and
images). This motivates us to propose a novel neural network policy (\domnet),
specifically designed to perform flexible relational reasoning over the tree-structured HTML
representation of websites.

We evaluate \emph{workflow-guided exploration} and \domnet
on a suite of web interaction tasks, including the MiniWoB benchmark of
\citep{shi2017wob}, the flight booking interface for Alaska Airlines, and a
new collection of tasks that we constructed to 
study additional challenges such as noisy environments, variation in
natural language, and longer time horizons.
Compared to previous results on MiniWoB \citet{shi2017wob}, which used 10
minutes of demonstrations per task (approximately 200 demonstrations on
average), our system achieves much higher success rates and establishes new
state-of-the-art results with only 3--10 demonstrations per task.

\section{Setup}

In the standard reinforcement learning setup,
an agent learns a policy $\pi(a|s)$ that
maps a state $s$ to a probability distribution 
over actions $a$.
At each time step $t$, 
the agent observes an environment state $s_t$ and chooses an action $a_t$,
which leads to a new state $s_{t+1}$ and a reward $r_t = r(s_t, a_t)$.
The goal is to maximize the expected return $\E[R]$,
where $R = \sum_t \gamma^t r_{t+1}$ and $\gamma$ is a discount factor.
Typical reinforcement learning agents
learn through trial-and-error:
rolling out episodes $(s_1, a_1, \dots, s_T, a_T)$
and adjusting their policy based on the results of those episodes.

We focus on settings where the reward is delayed and sparse.
Specifically, we assume that (1) the agent receives reward only at
the end of the episode, and (2) the reward is high (e.g., $+1$)
for only a small fraction of possible trajectories and is uniformly low (e.g., $-1$) otherwise.
With large state and action spaces,
it is difficult for the exploration policy to find episodes with positive rewards,
which prevents the policy from learning effectively.

We further assume that the agent is given a goal $g$,
which can either be
a structured key-value mapping
(e.g., \{task: forward, from: Bob, to: Alice\}) or
a natural language utterance
(e.g., \emph{``Forward Bob's message to Alice''}).
The agent's state $s$ consists of the goal $g$ and the current state of the web page,
represented as a tree of elements (henceforth \emph{DOM tree}).
We restrict the action space to click actions \texttt{Click(e)}
and type actions \texttt{Type(e,t)},
where \texttt{e} is a leaf element of the DOM tree,
and \texttt{t} is a string from the goal $g$ (a value from a structured goal,
or consecutive tokens from a natural language goal).
\reffig{workflow-lattice} shows an example episode for an email processing task.
The agent receives $+1$ reward if the task is completed correctly,
and $-1$ reward otherwise.

\section{Inducing workflows from demonstrations}\label{sec:workflows}

Given a collection of expert demonstrations $d = (\D s_1, \D a_1, \dots, \D s_T, \D a_T)$,
we would like explore actions $a_t$ that are ``similar'' to the demonstrated actions $\D a_t$.
Workflows capture this notion of similarity
by specifying a set of similar actions at each time step.
Formally, a workflow $z_{1:T}$ is a sequence of workflow steps,
where each step $z_t$
is a function that takes a state $s_t$ and returns
a constrained set $z_t(s_t)$ of similar actions.
We use a simple compositional constraint language (\refapp{constraint-language}) to describe workflow steps.
For example,
with $z_t = \texttt{Click(Tag("img"))}$,
the set $z_t(s_t)$ contains click actions on any DOM element in $s_t$ with tag \texttt{img}.

\begin{figure}\center
\includegraphics[width=\columnwidth]{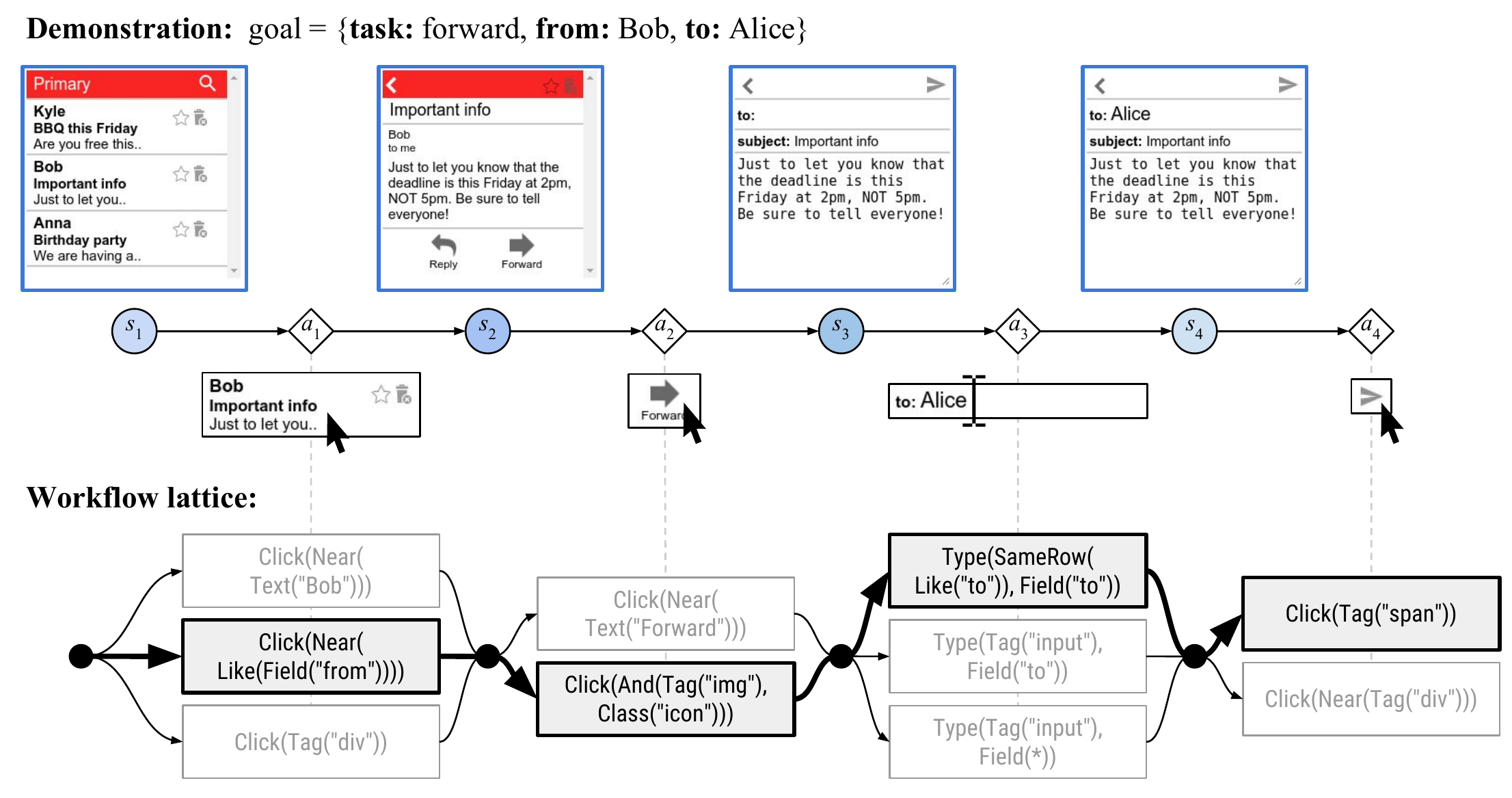}
\caption{
From each demonstration, we induce a workflow lattice
based on the actions in that demonstration.
Given a new environment, the workflow policy samples a workflow
(a path in the lattice, as shown in bold) and then samples actions that fit
the steps of the workflow.
}\label{fig:workflow-lattice}
\end{figure}

We induce a set of workflows from each demonstration
$d = (\D s_1, \D a_1, \dots, \D s_T, \D a_T)$ as follows.
For each time step $t$, we enumerate a set $Z_t$ of all possible workflow steps $z_t$
such that $\D a_t \in z_t(\D s_t)$. The set of workflows
is then the cross product $Z_1 \times \dots \times Z_T$ of the steps.
We can represent the induced workflows as paths
in a \emph{workflow lattice} as illustrated in \reffig{workflow-lattice}.

To handle noisy demonstrations where some actions are unnecessary
(e.g., when the demonstrator accidentally clicks on the background),
we add shortcut steps
that skip certain time steps.
We also add shortcut steps for any consecutive actions
that can be collapsed into a single equivalent action
(e.g., collapsing two type actions on the same DOM element
into a single \texttt{Type} step).
These shortcuts allow the lengths of the induced workflows to differ from the length of the demonstration.
We henceforth ignore these shortcut steps to simplify the notation.

The induced workflow steps are not equally effective.
For example in \reffig{workflow-lattice},
the workflow step \texttt{Click(Near(Text("Bob")))} (Click an element near text ``Bob'')
is too specific to the demonstration scenario,
while \texttt{Click(Tag("div"))} (Click on any \texttt{<div>} element)
is too general and covers too many irrelevant actions.
The next section describes how the workflow policy $\piw$
learns which workflow steps to use.

\section{Workflow exploration policy}\label{sec:workflow-policy}

Our workflow policy interacts with the environment to generate an episode in the following manner.
At the beginning of the episode,
the policy conditions on the provided goal $g$,
and selects a demonstration $d$ that carried out a similar goal:
\begin{equation}
  d \sim p(d|g) \propto \exp[\text{sim}(g, g_d)]
\end{equation}
where $\text{sim}(g,g_d)$ measures the similarity between $g$ and the goal $g_d$ of demonstration $d$.
In our tasks, we simply let $\text{sim}(g,g_d)$ be 1 if the structured
goals share the same keys, and $-\infty$ otherwise.

Then, at each time step $t$ with environment state $s_t$, we sample a workflow step $z_t$
according to the following distribution:
\begin{equation}
z_t \sim \piw(z | d,t) \propto \exp(\psi_{z,t,d}),
\end{equation}
where each $\psi_{z,t,d}$ is a separate scalar parameter to be learned.
Finally, we sample an action $a_t$ uniformly from the set $z_t(s_t)$.
\begin{equation}
a_t \sim p(a | z_t, s_t) = \frac{1}{|z_t(s_t)|}
\end{equation}

The overall probability of exploring an episode $e = (s_1, a_1, \dots, s_T, a_T)$ is then:
\begin{equation}
p(e | g) = p(d | g) \prod_{t=1}^{T} p(s_t | s_{t-1}, a_{t-1}) \sum_{z} p(a_t | z, s_t) \piw(z | d,t)
\end{equation}
where $p(s_t | s_{t-1}, a_{t-1})$ is the (unknown) state transition probability.

Note that $\piw(z | d, t)$ is not a function of the environment states $s_t$ at all. Its decisions
only depend on the selected demonstration and the current time $t$.
This \emph{environment-blindness}
means that the workflow policy uses far fewer parameters than a
state-dependent policy, enabling it to learn more quickly and preventing
overfitting. Due to \emph{environment-blindness}, the workflow policy cannot
solve the task, but it quickly learns to certain good behaviors, which
can help the neural policy learn.

To train the workflow policy, we use a variant of the REINFORCE algorithm \citep{williams1992simple,sutton1998reinforcement}.
In particular, after rolling out an episode $e = (s_1, a_1, \dots, s_T, a_T)$,
we approximate the gradient using the unbiased estimate
\begin{equation}
\sum_t (G_t - v_{d,t})\nabla_\psi \log \sum_{z} p(a_t | z, s_t) \piw(z | d,t),
\end{equation}
where $G_t$ is the return at time step $t$ and $v_{d,t}$ is a baseline term for variance reduction.

Sampled episodes from the workflow policy that receive a positive reward
are stored in a replay buffer, which will be used for training the neural policy $\pin$.

\section{Neural policy}\label{sec:neural-policy}

As outlined in \reffig{approach-overview},
the neural policy is learned using both on-policy and off-policy updates (where episodes are drawn from the replay buffer).
Both updates use A2C, the synchronous version of the advantage actor-critic
algorithm \citep{mnih2016asynchronous}.
Since only episodes with reward +1 enter the replay buffer, the off-policy updates
behave similarly to supervised learning on optimal trajectories.
Furthermore, successful episodes discovered during on-policy exploration are also added
to the replay buffer.

\paragraph{Model architecture.}
We propose \domnet, a neural architecture that captures the spatial and hierarchical 
structure of the DOM tree.
As illustrated in \reffig{neural-architecture},
the model first embeds the DOM elements and the input goal,
and then applies a series of attentions on the embeddings to finally produce a
distribution over actions $\pin(a|s)$ and a value function $V(s)$, the critic.
We highlight our novel DOM embedder,
and defer other details to \refapp{architecture-details}.

We design our DOM embedder to capture the various interactions
between DOM elements, similar to recent work in graph embeddings
\citep{kipf2017semi,pham2017column,hamilton2017inductive}.
In particular, DOM elements that are ``related''
(e.g., a checkbox and its associated label)
should pass their information to each other.

To embed a DOM element $e$,
we first compute the \emph{base embedding} $v_\mathrm{base}^e$
by embedding and concatenating its attributes
(tag, classes, text, etc.).
In order to capture the relationships between DOM elements, we next compute two 
types of \emph{neighbor embeddings}:

\begin{enumerate}
\item We define \emph{spatial neighbors} of $e$ to be
any element $e'$ within 30 pixels from $e$,
and then sum up their base embeddings to get the
\emph{spatial neighbor embedding} $v_\mathrm{spatial}^e$.
\item We define \emph{depth-$k$ tree neighbors} of $e$ to be
any element $e'$ such that the least common ancestor of $e$ and $e'$ in the DOM tree
has depth at most $k$.
Intuitively, tree neighbors of a higher depth are more related.
For each depth $k$, we apply a learnable affine transformation $f$ on the base embedding of
each depth-$k$ tree neighbor $e'$, and then apply max pooling to get
$v_{\mathrm{tree}[k]}^e = \max f(v_\mathrm{base}^{e'})$.
We let the \emph{tree neighbor embedding} $v_\mathrm{tree}^e$ be the
concatenation of $v_{\mathrm{tree}[k]}^e$ for $k = 3, 4, 5, 6$.
\end{enumerate}

Finally, we define the \emph{goal matching embedding} $v_\mathrm{match}^e$
to be the sum of the embeddings of all words in $e$ that also appear in the goal.
The final embedding $v_\mathrm{DOM}^e$ of $e$ is the concatenation
of the four embeddings $[v_\mathrm{base}^e; v_\mathrm{spatial}^e; v_\mathrm{tree}^e; v_\mathrm{match}^e]$.

\section{Experiments}
\subsection{Task setups}

We evaluate our approach on three suites of interactive web tasks:
\begin{enumerate}
\item \emph{MiniWoB}: the MiniWoB benchmark of \citet{shi2017wob}
\item \emph{MiniWoB++}: a new set of tasks that we constructed
to incorporate additional challenges not present in
MiniWoB, such as stochastic environments and variation in natural language.
\item \emph{Alaska}: the mobile flight booking interface for Alaska Airlines,
inspired by the FormWoB benchmark of \citet{shi2017wob}.
\end{enumerate}

We describe the common task settings of the MiniWoB and MiniWoB++ benchmarks,
and defer the description of the Alaska benchmark to \refsec{alaska}.

\paragraph{Environment.}
Each task contains a 160px $\times$ 210px
environment and a goal specified in text.
The majority of the tasks return a single sparse reward at
the end of the episode; either $+1$ (success) or $-1$ (failure). For greater
consistency among tasks, we disabled \emph{all} partial rewards in our
experiments.
The agent has access to the environment via a Selenium web driver interface.

The public MiniWoB benchmark\footnote{\url{http://alpha.openai.com/miniwob/}}
contains 80 tasks. We filtered for the 40 tasks that 
only require actions in our action space,
namely clicking on DOM elements and typing strings from the input goal.
Many of the excluded tasks involve somewhat specialized reasoning,
such as being able to compute the angle between two lines, or solve algebra problems.
For each task,
we used Amazon Mechanical Turk to collect 10 demonstrations,
which record all mouse and keyboard events along with
the state of the DOM when each event occurred.

\paragraph{Evaluation metric.}
We report \emph{success rate}: the percentage of test episodes with reward $+1$.
Since we have removed partial rewards, success rate is a linear scaling of the average reward,
and is equivalent to the definition of success rate in \citet{shi2017wob}.

\subsection{Main results} \label{sec:main-results}

\newcommand{\Sshi}{\textsc{Shi17}\xspace}
\newcommand{\Sbcrl}{\textsc{DOMnet+BC+RL}\xspace}
\newcommand{\Swge}{\textsc{DOMnet+WGE}\xspace}

\begin{figure}\center
\includegraphics[width=\textwidth]{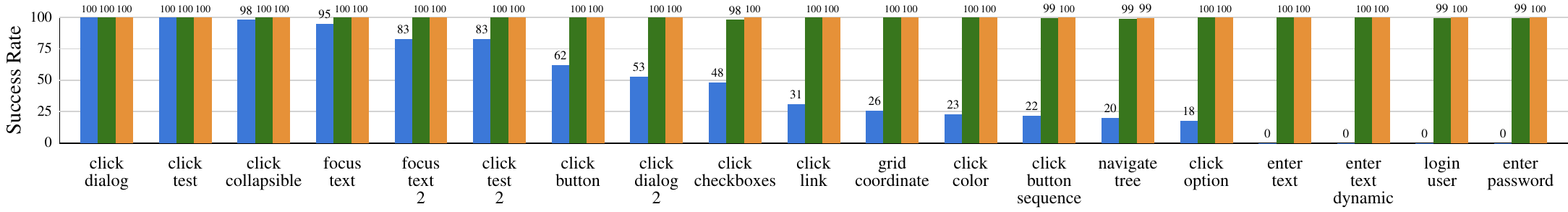}
\includegraphics[width=\textwidth]{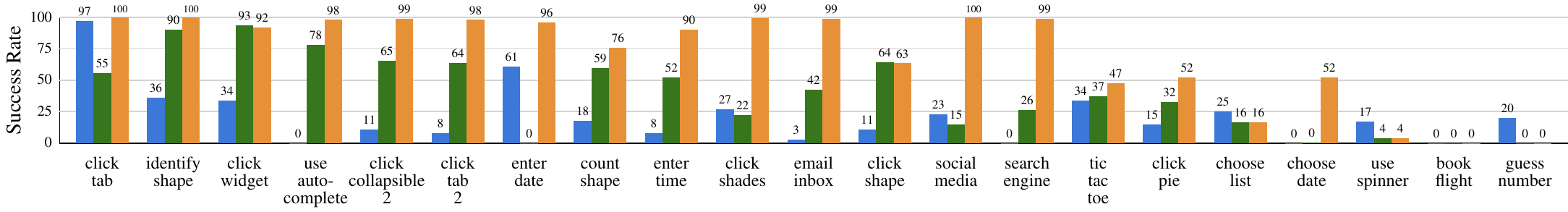}
\includegraphics[scale=.8]{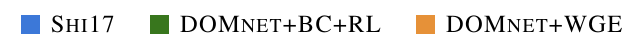}
\caption{
Success rates of different approaches
on the MiniWoB tasks. \Swge outperforms \Sshi on all but two tasks and
effectively solves a vast majority.
}\label{fig:main-results-chart}
\end{figure}

\begin{table}\center\small
\begin{tabular}{@{}ll@{}r|rrr}
\textbf{Task} & \textbf{Description} & \textbf{Steps} & \textbf{BC+RL} &
\textbf{$\piw$ only} & \textbf{WGE} \\ \hline
click-checkboxes & Click 0--6 specified checkboxes & 7  & 98 & 81 & \textbf{100} \\
click-checkboxes-large$^+$ & \dots 5--12 targets   & 13 & 0 & 43 & \textbf{84} \\
click-checkboxes-soft$^+$ & \dots specifies synonyms of the targets & 7 & 51 & 34& \textbf{94} \\
click-checkboxes-transfer$^+$ & \dots training data has 0-3 targets & 7 & \textbf{64} & 17 & \textbf{64} \\
multi-ordering$^+$ & Fill a form with varying field orderings & 4 & 5 & 78 & \textbf{100} \\
multi-layout$^+$ & Fill a form with varying UIs layouts & 4 & 99 & 9 & \textbf{100} \\
social-media & Do an action on the specified Tweet & 2 & 15 & 2 & \textbf{100} \\
social-media-all$^+$ & \dots on all matching Tweets & 12 & \textbf{1} & 0
& 0 \\
social-media-some$^+$ & \dots on specified no. of matching Tweets & 12 & 2 &
 3 & \textbf{42} \\ \hline
email-inbox & Perform tasks on an email inbox & 4 & 43 & 3 & \textbf{99} \\
email-inbox-nl$^+$ & \dots natural language goal & 4 & 28 & 0 & \textbf{93} \\
\end{tabular}
\caption{Results on additional tasks. ($+$ = MiniWoB++,
Steps = task length as the maximum number of steps needed for a perfect policy to complete the task)
}
\label{tab:analysis}
\end{table}

We compare the success rates across the MiniWoB tasks of the following approaches:
\begin{itemize}
\item \Sshi: the system from \citet{shi2017wob}, pre-trained
with behavioral cloning on 10 minutes of demonstrations (approximately 200
demonstrations on average) and fine-tuned with RL. Unlike \domnet, this system
primarily uses a pixel-based representation of the state.\footnote{It
is augmented with filters that activate on textual elements which overlap with
goal text.}
\item \Sbcrl: our proposed neural policy, \domnet, but
pre-trained with behavioral cloning on 10 demonstrations
and fine-tuned with RL, like \Sshi.
During behavioral cloning,
we apply early stopping based on the reward on a validation set.
\item \Swge: our proposed neural policy, \domnet, trained with
workflow-guided exploration on 10 demonstrations.
\end{itemize}

For \Sbcrl and \Swge, we report the test success rate at the time step where
the success rate on a validation set reaches its maximum.

The results are shown in \reffig{main-results-chart}.
By comparing \Sshi with \Sbcrl, we can roughly evaluate the contribution
of our new neural architecture \domnet, since the two share the same training procedure (BC+RL).
While \Sshi also uses the DOM tree to compute text alignment features
in addition to the pixel-level input,
our \domnet uses the DOM structure more explicitly.
We find \Sbcrl to empirically improve the success rate over \Sshi on most tasks.

By comparing \Sbcrl and \Swge, we find that
workflow-guided exploration enables \domnet to
perform even better on the more difficult tasks, which we analyze in the next section.
Some of the workflows that the workflow policy $\piw$ learns
are shown in \refapp{example-workflows}.

\subsection{Analysis}\label{sec:analysis}

\subsubsection{MiniWoB++ benchmark}\label{sec:plusplus}

We constructed and released the MiniWoB++ benchmark of tasks to
study additional challenges a web agent might encounter, including:
longer time horizons (click-checkboxes-large),
``soft'' reasoning about natural language (click-checkboxes-soft),
and stochastically varying layouts (multi-orderings, multi-layouts).
\reftab{analysis} lists the tasks and their time horizons (number of steps
needed for a perfect policy to carry out the longest goal) as a crude measure
of task complexity.

We first compare the performance of \domnet trained with BC+RL (baseline)
and \domnet trained with WGE (our full approach).
The proposed WGE model outperforms the BC+RL model by an average of 42\%
absolute success rate.
We analyzed their behaviors and noticed two common failure modes of training
with BC+RL that are
mitigated by instead training with WGE:
\begin{enumerate}
\item The BC+RL model has a tendency to take actions that prematurely
terminate the episode (e.g., hitting ``Submit'' in click-checkboxes-large
before all required boxes are checked).
One likely cause is that these actions occur across all
demonstrations, while other non-terminating actions
(e.g., clicking different checkboxes)
vary across demonstrations.
\item The BC+RL model occasionally
gets stuck in cyclic behavior such as repeatedly checking and unchecking the
same checkbox. These failure modes stem from overfitting to parts of the
demonstrations, which WGE avoids.
\end{enumerate}

Next, we analyze the workflow policy $\piw$ learned by WGE.
The workflow policy $\piw$ by itself is too simplistic to work well at test time for several reasons:
\begin{enumerate}
\item Workflows ignore environment state and therefore cannot respond to the differences
in the environment, such as the different layouts in multi-layouts.
\item The workflow constraint language lacks the expressivity to specify certain
actions, such as clicking on synonyms of a particular word in click-checkboxes-soft.
\item The workflow policy lacks expressivity to select the correct workflow for a given goal.
\end{enumerate}

Nonetheless the workflow policy $\piw$ is sufficiently constrained to discover reward some
of the time, and the neural policy $\pin$ is able to learn the right behavior from such
episodes.
As such, the neural policy can achieve high success rates
even when the workflow policy $\piw$ performs poorly.

\subsubsection{Natural language inputs}\label{sec:nlp}

While MiniWoB tasks provide structured goals,
we can also apply our approach to natural language goals.
We collected a training dataset using the overnight data collection technique \citep{wang2015overnight}.
In the email-inbox-nl task,
we collected natural language templates
by asking annotators to paraphrase the task goals
(e.g., \emph{``Forward Bob's message to Alice''} $\to$
\emph{``Email Alice the email I got from Bob''})
and then abstracting out the fields
(\emph{``Email \texttt{<TO>} the email I got from \texttt{<FROM>}''}).
During training, the workflow policy $\piw$
receives states with both the structured goal
and the natural language utterance generated from a random template,
while the neural policy $\pin$ receives only the utterance.
At test time, the neural policy is evaluated on unseen utterances.
The results in \reftab{analysis} show that 
the WGE model can learn to understand natural language goals (93\% success rate).

Note that the workflow policy needs access to the structured inputs
only because our constraint language for workflow steps operates on structured inputs.
The constraint language
could potentially be modified to work with utterances directly
(e.g., \texttt{After("to")} extracts the utterance word after \emph{``to''}),
but we leave this for future work.

\subsubsection{Scaling to real world tasks}\label{sec:alaska}

We applied our approach on the Alaska benchmark,
a more realistic flight search task on the Alaska Airlines mobile site
inspired by the FormWoB task in \citet{shi2017wob}.
In this task, the agent must complete the flight search form
with the provided information (6--7 fields).
We ported the web page to the MiniWoB framework
with a larger 375px $\times$ 667px screen,
replaced the server backend with
a surrogate JavaScript function, and clamped the environment date to March 1, 2017.

Following \citet{shi2017wob},
we give partial reward based on the fraction of correct fields in the submitted form
if all required fields are filled in.
Despite this partial reward, the reward is still extremely sparse:
there are over 200 DOM elements (compared to $\approx$ 10--50 in MiniWoB tasks),
and a typical episode requires at least 11 actions
involving various types of widgets such as autocompletes and date pickers.
The probability that a random agent gets positive
reward is less than $10^{-20}$.

We first performed experiments on Alaska-Shi17, a clone of the original Alaska
Airlines task in \citet{shi2017wob}, where the goal always specifies
a roundtrip flight (two airports and two dates).
On their dataset,
our approach, using only 1 demonstration, achieves an average reward of 0.97,
compared to their best result of 0.57, which uses around 80 demonstrations.

Our success motivated us to test on a more difficult version of the task which
additionally requires selecting flight type (a checkbox for one-way flight),
number of passengers (an increment-decrement counter),
and seat type (hidden under an accordion).
We achieve an average reward of 0.86
using 10 demonstrations. This demonstrates our method can handle long horizons
on real-world websites.

\subsubsection{Sample efficiency}\label{sec:sample}

\begin{figure}\center
\includegraphics[width=\textwidth]{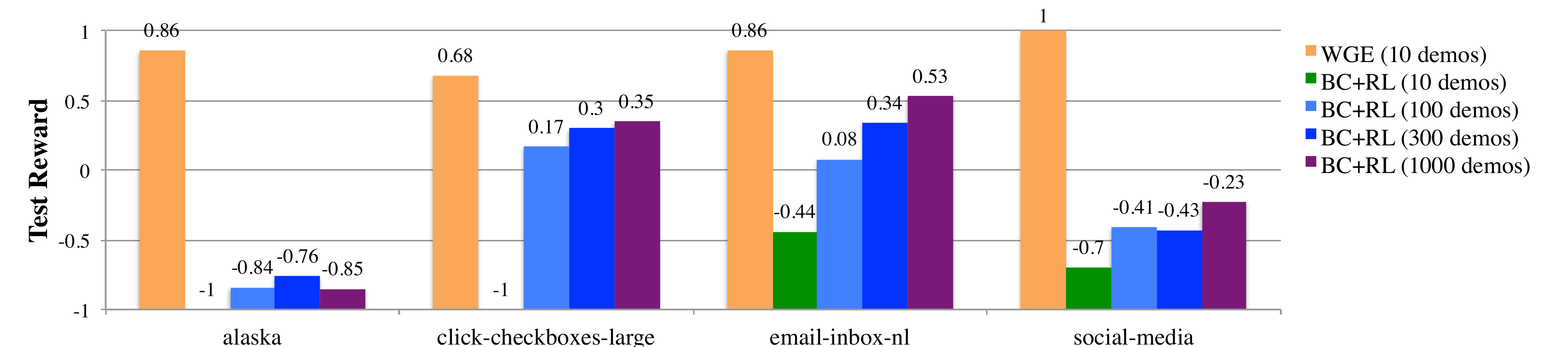}
\caption{
Comparison between \Sbcrl and \Swge on several of the most difficult tasks,
evaluated on test reward. \Swge trained on 10 demonstrations outperforms
\Sbcrl even with 1000 demonstrations.
}\label{fig:sample_efficiency}
\end{figure}

To evaluate the demonstration efficiency of our approach, we compare \Swge
with \Sbcrl trained on increased numbers of demonstrations. We compare \Swge
trained on $10$ demonstrations with \Sbcrl on $10$, $100$, $300$, and $1000$
demonstrations. The test rewards\footnote{We report test
reward since success rate is artificially high in
the Alaska task due to partial rewards.} on several of the hardest tasks are
summarized in \reffig{sample_efficiency}.

Increasing the number of demonstrations improves the performance of BC+RL, as
it helps prevent overfitting. However, on every evaluated task, WGE trained
with only $10$ demonstrations still achieves much higher test reward than
BC+RL with $1000$ demonstrations. This corresponds to an over 100x sample
efficiency improvement of our method over behavioral cloning in terms of the
number of demonstrations.

\section{Discussion}

\paragraph{Learning agents for the web.}
Previous work on learning agents for web interactions falls into two main
categories. First, simple programs may be specified by the user \citep{yeh2009sikuli}
or may be inferred from demonstrations \citep{allen2007plow}. Second, soft
policies may be learned from scratch or ``warm-started'' from demonstrations
\citep{shi2017wob}. Notably, sparse rewards prevented \citet{shi2017wob} from
successfully learning, even when using a moderate number of demonstrations.
While policies have proven to be more difficult to learn, they have the
potential to be expressive and flexible. Our work takes a step in this
direction.

\paragraph{Sparse rewards without prior knowledge.}
Numerous works attempt to address sparse rewards without incorporating any
additional prior knowledge. Exploration methods \citep{osband2016deep,
chentanez2005intrinsically, weber2017imagination} help the agent better
explore the state space to encounter more reward; shaping rewards
\citep{ng1999policy} directly modify the reward function to encourage certain
behaviors; and other works \citep{jaderberg2016reinforcement,
andrychowicz2017hindsight} augment the reward signal with additional
unsupervised reward. However, without prior knowledge, helping the agent
receive additional reward is difficult in general.

\paragraph{Imitation learning.}
Various methods have been proposed to leverage additional signals from experts.
For instance,
when an expert policy is available, methods such as
\textsc{DAgger} \citep{ross2011reduction} and
\textsc{AggreVaTe} \citep{ross2014reinforce,sun2017deeply}
can query the expert policy
to augment the dataset for training the agent.
When only expert \emph{demonstrations} are available,
inverse reinforcement learning methods
\citep{abbeel2004apprenticeship,ziebart2008maximum,finn2016guided,ho2016generative,baram2017end}
infer a reward function from the demonstrations without using reinforcement signals from the environment.

The usual method for incorporating both demonstrations and reinforcement signals
is to pre-train the agent with demonstrations before applying RL.
Recent work extends this technique by
(1) introducing different objective functions and regularization during pre-training,
and (2) mixing demonstrations and rolled-out episodes during RL updates
\citep{hosu2016playing,hester2018deep,vecerik2017leveraging,nair2017overcoming}.

Instead of training the agent on demonstrations directly,
our work uses demonstrations to \emph{guide exploration}.  The core idea is to
explore trajectories that lie in a ``neighborhood'' surrounding an expert
demonstration.  In our case, the neighborhood is defined by a workflow, which
only permits action sequences analogous to the demonstrated actions.
Several previous works also explore neighborhoods of demonstrations via reward shaping
\citep{brys2015reinforcement, hussein2017deep} or off-policy sampling \citep{levine2013guided}.
One key distinction of our work is that we define neighborhoods in terms of action
similarity rather than state similarity. This distinction is particularly
important for the web tasks: we can easily and intuitively describe how two
actions are analogous (e.g., ``they both type a username into a textbox''),
while it is harder to decide if two web page states are analogous (e.g., the
email inboxes of two different users will have completely different emails, but
they could still be analogous, depending on the task.)

\paragraph{Hierarchical reinforcement learning.}
Hierarchical reinforcement learning (HRL) methods decompose complex tasks into
simpler subtasks that are easier to learn. Main HRL frameworks include
abstract actions \citep{sutton1999between, konidaris2007building,
hauser2008using}, abstract partial policies \citep{parr1998reinforcement},
and abstract states \citep{roderick2017deep, dietterich1998maxq,
li2006towards}. These frameworks require varying amounts of prior knowledge.
The original formulations required programmers to manually specify the
decomposition of the complex task, while \citet{andreas2016modular} only
requires supervision to identify subtasks, and \citet{bacon2017option,
daniel2016hierarchical} learn the decomposition fully automatically, at the
cost of performance.

Within the HRL methods, our work is closest to \citet{parr1998reinforcement}
and the line of work on constraints in robotics \citep{phillips2016learning,
perez2017c}.
The work in \citet{parr1998reinforcement} specifies partial policies, which
constrain the set of possible actions at each state, similar to our workflow
items. In contrast to previous instantiations of the HAM framework
\citep{andre2003programmable, marthi2005concurrent}, which require programmers to
specify these constraints manually, our work automatically induces
constraints from user demonstrations, which do not require special
skills to provide. \citet{phillips2016learning, perez2017c} also resemble our
work, in learning constraints from demonstrations, but differ in the way they
use the demonstrations. Whereas our work uses the learned constraints for
exploration, \citet{phillips2016learning} only uses the constraints for
planning and \citet{perez2017c} build a knowledge base of constraints to use
at test time.

\paragraph{Summary.}
Our workflow-guided framework represents a judicious combination of demonstrations, abstractions, and
expressive neural policies.
We leverage the targeted information of demonstrations and the inductive bias of workflows.
But this is only used for exploration, protecting the expressive neural policy from overfitting.
As a result, we are able to learn rather complex policies from a very sparse reward signal and very few demonstrations.

\paragraph{Acknowledgments.}
This work was supported by NSF CAREER Award IIS-1552635.

\paragraph{Reproducibility.}
Our code and data are available at \url{https://github.com/stanfordnlp/wge}.
Reproducible experiments are available on the CodaLab platform at
\url{https://worksheets.codalab.org/worksheets/0x0f25031bd42f4aabbc17625fe1484066/}.

\bibliography{all}
\bibliographystyle{iclr2018_conference}

\appendix
\clearpage
\section{Constraint language for workflow steps}\label{sec:constraint-language}

\newcommand\C[1]{{\it #1}}
\newcommand\T[1]{{\tt #1}}

We try to keep the constraint language as minimal and general as possible.  The
main part of the language is the object selector (\C{elementSet}) which selects
either (1) objects that share a specified property, or (2) objects that align
spatially.  These two types of constraints should be applicable in many typical
RL domains such as game playing and robot navigation.

\begin{center}
\renewcommand{\arraystretch}{1.2}
\begin{tabular}{|rcl|} \hline
\C{constraint}
& ::= & \T{Click(\C{elementSet})} \\
&& [Any click action on an element in \C{elementSet}] \\
& | & \T{Type(\C{elementSet},\C{string})} \\
&& [Any type action that types \C{string} on an element in \C{elementSet}] \\
& | & \T{Type(\C{elementSet},Field(*))} \\
&& [Any type action that types a goal field value on an element in \C{elementSet}] \\
\C{elementSet}
& ::= & \T{Tag(\C{tag})} \\
&& [Any element with HTML tag \C{tag}] \\
& | & \T{Text(\C{string})} \\
&& [Any element with text \C{string}] \\
& | & \T{Like(\C{string})} \\
&& [Any element whose text is a substring of \C{string}] \\
& | & \T{Near(\C{elementSet})} \\
&& [Any element that is within 30px from an element in \C{elementSet}] \\
& | & \T{SameRow(\C{elementSet})} \\
&& [Any element that aligns horizontally with an element in \C{elementSet}] \\
& | & \T{SameCol(\C{elementSet})} \\
&& [Any element that aligns vertically with an element in \C{elementSet}] \\
& | & \T{And(\C{elementSet},Class(\C{classes}))} \\
&& [Any element from \C{elementSet} matching some class name in \C{classes}] \\
\C{tag}
& ::= & a valid HTML tag name \\
\C{string}
& ::= & a string literal \\
& | & \T{Field(\C{fieldName})} \\
&& [The value from the goal field \C{fieldName}] \\
\C{classes}
& ::= & a list of valid HTML class names \\ \hline
\end{tabular}
\end{center}

To avoid combinatorial explosion of relatively useless constraints,
we limit the number of nested \C{elementSet} applications to 3,
where the third application must be the \T{Class} filter.
When we induce workflow steps from a demonstration,
the valid literal values for \C{tag}, \C{string}, and \C{classes}
are extracted from the demonstration state.





\section{Examples of learned workflows}\label{sec:example-workflows}

\begin{center}\small
\begin{tabular}{c}
\textbf{login-user} \\
\emph{Enter the username "ashlea" and password "k0UQp" and press login.} \\
\{username: ashlea, password: k0UQp\} \\ \hline
\texttt{Type(Tag("input\_text"),Field("username"))} \\
\texttt{Type(Tag("input\_password"),Field("password"))} \\
\texttt{Click(Like("Login"))} \\ \hline
\end{tabular}
\end{center}

\begin{center}\small
\begin{tabular}{c}
\textbf{email-inbox} \\
\emph{Find the email by Ilka and forward it to Krista.} \\
\{task: forward, name: Ilka, to: Krista\} \\ \hline
\texttt{Click(Near(Field("by")))} \\
\texttt{Click(SameCol(Like("Forward")))} \\
\texttt{Type(And(Near("Subject"),Class("forward-sender")),Field("to"))} \\
\texttt{Click(Tag("span"))} \\ \hline
\end{tabular}
\end{center}

\begin{center}\small
\begin{tabular}{c}
\textbf{search-engine} \\
\emph{Enter "Cheree" and press "Search", then find and click the 5th search result.} \\
\{target: Cheree, rank: 5\} \\ \hline
\texttt{Type(Near(Tag("button")),Field(*))} \\
\texttt{Click(Text("Search"))} \\
\texttt{Click(Like(">"))} \\
\texttt{Click(Text(Field("target")))} \\ \hline
\end{tabular}
\end{center}

\begin{center}\small
\begin{tabular}{c}
\textbf{Alaska} \\
\{departure city: Tampa, destination city: Seattle, ticket type: return flight, \\
departure day: 6, returning Day: 16, passengers: 3, seat type: first \} \\ \hline
\scriptsize\texttt{Type(And(Near(Like("From")),Class("text-input-pad")),Field("departure city"))} \\
\scriptsize\texttt{Click(And(SameRow(Tag("label")),Class(["input-selection","last"])))} \\
\scriptsize\texttt{Type(And(Near(Like("To")),Class("text-input-pad")),Field("destination city"))} \\
\scriptsize\texttt{Click(Like(Field("destination city")))} \\
\scriptsize\texttt{Click(And(SameCol(Tag("a")),Class(["calbg","text-input"])))} \\
\scriptsize\texttt{Click(Text(Field("departure day")))} \\
\scriptsize\texttt{Click(Like("Done"))} \\
\scriptsize\texttt{Click(Near(Like("Return")),Class(["calbg","text-input"]))} \\
\scriptsize\texttt{Click(Text(Field("returning day")))} \\
\scriptsize\texttt{Click(Like("Done"))} \\
\scriptsize\texttt{Click(Like("+"))} \\
\scriptsize\texttt{Click(Like("+"))} \\
\scriptsize\texttt{Click(Tag("h2"))} \\
\scriptsize\texttt{Click(Text("First"))} \\
\scriptsize\texttt{Click(And(Near(Tag("body")),Class("button")))} \\ \hline
\end{tabular}
\end{center}

\section{Details of the neural model architecture}\label{sec:architecture-details}

\begin{figure}[tb]\center
\includegraphics[height=0.65\textheight]{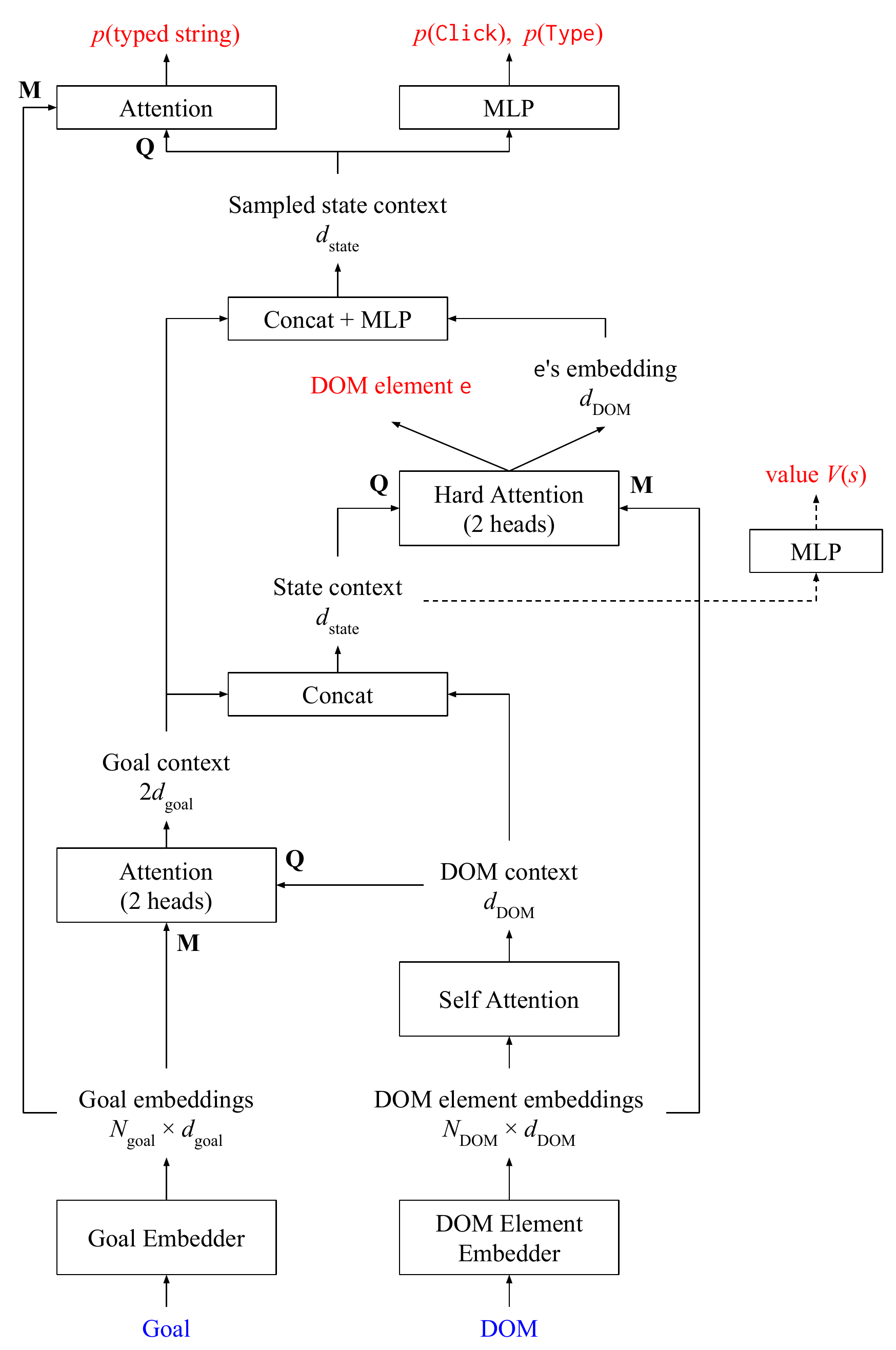}
\caption{
The architecture of the neural policy $\pin$.
The inputs from the state are denoted in blue, while the outputs are denoted in red.
\textbf{Q} = query vector; \textbf{M} = memory matrix.
}\label{fig:neural-architecture}
\end{figure}

\paragraph{Embeddings.}

From the input state,
we first embed the DOM elements $e$ and the goal units $u$,
where $u$ is a key-value pair for structured goals
and a token for natural language goals.

The process for computing the embedding $v_\mathrm{DOM}^e$ of DOM elements
is already described in \refsec{neural-policy}.
For the goal unit embedding $v_\mathrm{goal}^u$,
we embed each key-value pair as the sum of word embeddings,
and embed natural language goals with an LSTM.

\paragraph{Attentions.}

After obtaining the embedding $v_\mathrm{DOM}^e$ of each DOM element $e$
and $v_\mathrm{goal}^u$ of each goal unit $u$,
we apply a series of attentions to relate the DOM elements
with the goal:

\begin{enumerate}
\item
\emph{DOM context}: we applied max-pooling on $v_\mathrm{DOM}^e$ to get a query vector,
and then attend over the DOM embeddings $v_\mathrm{DOM}^e$. The DOM context is the weighted average
of the attended DOM embeddings.
\item
\emph{Goal contexts}: we use the DOM context as the query vector
to attend over the goal embeddings $v_\mathrm{goal}^u$. We compute two goal contexts
from two different attention heads. Each head uses sentinel attention,
where part of the attention can be put on a learned NULL vector,
which is useful for ignoring the goal when the next action should not depend
on the goal.
\item
\emph{DOM element selection}: We concatenate the DOM context and
goal contexts into a query vector to attend over on the DOM embeddings $v_\mathrm{DOM}^e$.
We use two attention heads, and combine the attention weights from the two heads
based on ratio computed from the goal contexts.
The result is a distribution over the target DOM elements \texttt{e}.
\item
\emph{Typed string and action selection}:
For a given target DOM element \texttt{e}, we combine the goal context
and the embedding $v_\mathrm{DOM}^\texttt{e}$ of \texttt{e} to get a query vector
to attend over the goal embeddings $v_\mathrm{goal}^u$.
For structured queries, we get a distribution over the goal fields,
while for natural language queries, we get distributions of the
start and end tokens.
The same query vector is also used to compute the distribution
over the action types (\texttt{click} or \texttt{type}).
\end{enumerate}

\end{document}